\documentclass[runningheads]{llncs}

\usepackage{graphicx}
\usepackage{amsmath}
\usepackage{amssymb}
\usepackage{algorithm}
\usepackage[noend]{algorithmic}

\usepackage{verbatim}
\renewcommand*{\vec}[1]{\ensuremath{\boldsymbol{#1}}}
\newcommand*{\Matrix}[1]{\ensuremath{\boldsymbol{#1}}}

\begin{document}

\title{The Temporal Dictionary Ensemble (TDE) Classifier for Time Series Classification}

\titlerunning{The TDE Classifier for Time Series Classification}

\author{Matthew Middlehurst \and James Large \and Gavin Cawley \and Anthony Bagnall}

\authorrunning{Middlehurst et al.}

\institute{
School of Computing Sciences, University of East Anglia, Norwich, UK\\
\email{M.Middlehurst@uea.ac.uk}
}

\maketitle

\begin{abstract}
Using bag of words representations of time series is a popular approach to time series classification. These algorithms involve approximating and discretising windows over a series to form words, then forming a count of words over a given dictionary. Classifiers are constructed on the resulting histograms of word counts. A 2017 evaluation of a range of time series classifiers found the bag of symbolic-fourier approximation symbols (BOSS) ensemble the best of the dictionary based classifiers. It forms one of the components of hierarchical vote collective of transformation-based ensembles (HIVE-COTE), which represents the current state of the art. Since then, several new dictionary based algorithms have been proposed that are more accurate or more scalable (or both) than BOSS. We propose a further extension of these dictionary based classifiers that combines the best elements of the others combined with a novel approach to constructing ensemble members based on an adaptive Gaussian process model of the parameter space. We demonstrate that the temporal dictionary ensemble (TDE) is more accurate  than other dictionary based approaches. Furthermore, unlike the other classifiers, if we replace BOSS in HIVE-COTE with TDE, HIVE-COTE is significantly more accurate. We also show this new version of HIVE-COTE is significantly more accurate than the current best deep learning approach, a recently proposed hybrid tree ensemble and a recently introduced competitive classifier making use of highly randomised convolutional kernels. This advance represents a new state of the art for time series classification.

\end{abstract}

\section{Introduction}
\label{sec:intro}

    Dictionary based approaches adapt the bag of words model commonly used in signal processing, computer vision and audio processing for time series classification (TSC).
    A comparison of TSC algorithms, commonly known as the bake off, formed a taxonomy of approaches based on representations of discriminatory features, with dictionary approaches being one of these.
    From the bake off the bag of Symbolic-Fourier Approximation symbols (BOSS)~\cite{schafer2015boss} ensemble was found to be the most accurate dictionary classifier by a significant amount.
    BOSS was found to be the third most accurate algorithm out of the 20 compared. This highlights the utility of dictionary methods for TSC.
    
    This performance lead to BOSS being incorporated into the hierarchical vote collective of transformation-based ensembles (HIVE-COTE)~\cite{lines2018time}, a heterogeneous ensemble encompassing multiple representations.
    The inclusion of BOSS and the subsequent significant improvement in accuracy places HIVE-COTE in the state of the art for TSC among three other algorithms proposed more recently.
    These are the time series combination of heterogeneous and integrated embeddings forest (TS-CHIEF)~\cite{shifaz2020ts}, which also a hybrid of multiple representations, the random convolutional kernel transform (ROCKET)~\cite{dempster2019rocket}, and the deep learning approach InceptionTime~\cite{fawaz2019inceptiontime}.
    
    Since the bake off a number of dictionary algorithms have been published, focusing on improving accuracy~\cite{schafer2017fast,large2019time}, prediction time efficiency~\cite{schafer2017fast}, train time and memory efficiency~\cite{middlehurst2019scalable}.
    These algorithms are mostly extensions of BOSS, making alterations to different parts of the original algorithm.
    Word extraction for time series classification (WEASEL)~\cite{schafer2017fast} abandons the ensemble structure in favour of feature selection and changes the method of word discretisation.
    Spatial BOSS (S-BOSS)~\cite{large2019time} introduces temporal information and additional features using spatial pyramids.
    Contractable BOSS (cBOSS)~\cite{middlehurst2019scalable} changes the method used by BOSS to form its ensemble to improve efficiency and allow for a number of usability improvements.
    
    Each of these methods constitutes an improvement to the dictionary representation from BOSS. Our contribution is to combine design features of these four classifiers (BOSS, WEASEL, S-BOSS and cBOSS) to make a new algorithm, the Temporal Dictionary Ensemble (TDE). Like BOSS, TDE is a homogeneous ensemble of nearest neighbour classifiers that use distance between histograms of word counts and injects diversity through parameter variation. TDE takes the ensemble structure from cBOSS, which is more robust and scaleable. The use of spatial pyramids is adapted from S-BOSS. WEASEL uses bi-grams and an alternative method of finding word breakpoints. This too is employed by TDE. 
    
    We found the simplest way of combining these components did not result in significant improvement. We speculate that the massive increase in the parameter space made the randomised diversity mechanism result in too many poor learners in the ensemble. We propose a novel mechanism of base classifier model selection based on an adaptive form of Gaussian process (GP) modelling of the parameter space. Through extensive evaluation with the UCR time series classification repository~\cite{dau2019ucr}, we show that  TDE is significantly more accurate than WEASEL and S-BOSS while retaining the usability and scalability of cBOSS. We further show that if TDE replaces BOSS in HIVE-COTE, the resulting classifier is significantly more accurate than HIVE-COTE with BOSS and all three competing state of the art classifiers.
    
    The rest of this paper is structured as follows. 
    Section~\ref{sec:background} provides background information for the four dictionary based algorithms relevant to TDE. 
    Section~\ref{sec:tde} describes the TDE algorithm, including the GP based parameter search. Section~\ref{sec:results} presents the performance evaluation of TDE. 
    Conclusions are drawn in Section~\ref{sec:conclusion} and future work is discussed.
    
\section{Dictionary Based Classifiers}
\label{sec:background}

Dictionary based classifiers have the same broad structure. A sliding window of length $w$ is run across a series. For each window, the real valued series of length $w$ is converted through approximation and discretisation processes into a symbolic string of length $l$, which consists of $\alpha$ possible letters. The occurrence in a series of each `word' from the dictionary defined by $l$ and $\alpha$ is counted, and once the sliding window has completed the series is transformed into a histogram. Classification is based on the histograms of the words extracted from the series, rather than the raw data.  

The Bags of Symbolic-Fourier-Approximation Symbols (BOSS)~\cite{schafer2015boss} was found to be the most accurate dictionary based classifier in a 2017 study~\cite{bagnall2017great}. Hence, it forms our benchmark for new dictionary based approaches. BOSS is described in detail in Section~\ref{sec:boss}. A number of extensions and alternatives to BOSS have been proposed.

\begin{itemize}
    \item One of the problems with BOSS is that it can be memory and time inefficient, especially on data where many transforms are accepted into the final ensemble. cBOSS (Section~\ref{sec:cboss}) addresses the scalability issues of BOSS~\cite{middlehurst2019scalable} by altering the ensemble structure.
    \item BOSS ignores the temporal location of patterns. Rectifying this led to an extension of BOSS based on spatial pyramids, called S-BOSS~\cite{large2019time}, described in Section~\ref{sec:sboss}.
    \item WEASEL~\cite{schafer2017fast} is a dictionary based classifier by the same team that produced BOSS. It is based on feature selection from histograms for a linear model (see Section~\ref{sec:weasel}).
\end{itemize}

We propose a dictionary classifier that merges these extensions and improvements to the core concept of BOSS, called the Temporal Dictionary Ensemble (TDE). It lends from the sped-up ensemble structure of cBOSS, the spatial pyramid structure of S-BOSS, and the word and histogram forming improvements of WEASEL. 
TDE is fully described in Section~\ref{sec:tde}.

\subsection{Bags of Symbolic-Fourier-Approximation Symbols (BOSS)~\cite{schafer2015boss}}
\label{sec:boss}

Algorithm~\ref{alg:boss} gives a formal description of the bag forming process of an individual BOSS classifier. Words are created using Symbolic Fourer Approximation (SFA)~\cite{schafer2012sfa}. SFA first finds the Fourier transform of the window (line 8), then discretises the first $l$ Fourier terms into $\alpha$ symbols to form a word, using a bespoke supervised discretisation algorithm called Multiple Coefficient Binning (MCB) (line 13). It has an option to normalise each window or not by dropping the first Fourier term (lines 6-7). Lines 14-16 encapsulates the process of not counting trivially self similar words: if two consecutive windows produce the same word, the second occurrence is ignored. This is to avoid a slow-changing pattern relative to the window size being over-represented in the resulting histogram.

BOSS uses a non-symmetric distance function in conjunction with a nearest neighbour classifier. Only the words contained in the test instance's histogram (i.e. the word count is above zero) are used in the distance calculation, but it is otherwise the Euclidean distance. 

    \begin{algorithm}[ht]
    	\caption{baseBOSS(A list of $n$ time series of length $m$, ${\bf T}=({\bf X,y})$)}
    	\label{alg:boss}
    	\begin{algorithmic}[1]
    \REQUIRE the word length $l$, the alphabet size $\alpha$, the window length $w$, normalisation parameter $p$
    		\STATE Let ${\bf H}$ be a list of $n$ histograms $({\bf h}_1,\ldots,{\bf h}_n)$
    		\STATE Let ${\bf B}$ be a matrix of $l$ by $\alpha$ breakpoints found by MCB
    		\FOR {$i \leftarrow  1$ to $n$}
    			\FOR {$j \leftarrow 1$ to $m-w+1$}
    				\STATE ${\bf s}\leftarrow x_{i,j} \ldots x_{i,j+w-1}$
    				\IF{$p$}
    				    \STATE $s \leftarrow $normalise($s$)
    				\ENDIF
    				\STATE ${\bf q} \leftarrow$ DFT(${\bf s}, l, \alpha$,$p$) \COMMENT{ {\em {\bf q} is a vector of the complex DFT coefficients}}
    				\IF{$p$}
        			    \STATE ${\bf q'} \leftarrow (q_2 \ldots q_{l/2+1})$
        			\ELSE
        			    \STATE ${\bf q'} \leftarrow (q_1 \ldots q_{l/2})$
        			\ENDIF
    				\STATE ${\bf r} \leftarrow$ SFAlookup(${\bf q', B}$)
    				\IF{${\bf r} \neq {\bf p}$}
    					\STATE $pos \leftarrow $index(${\bf r}$)
    					\STATE ${h}_{i,pos} \leftarrow {h}_{i,pos} + 1$
    				\ENDIF
    				\STATE ${\bf p} \leftarrow {\bf r} $
    			\ENDFOR
    		\ENDFOR
    	\end{algorithmic}
    \end{algorithm}
    
The final classifier is an ensemble of individual BOSS classifiers (parameterised transform plus nearest neighbour classifier) found through first fitting and evaluating a large number of individual classifiers, then retaining only those within 92\% accuracy of the best classifier. The BOSS ensemble (also referred to as just BOSS), evaluates and retains the best of all transforms parameterised in the range $ w \in \{10 \ldots m\}$ with $m/4$ values where m is the length of the series, $l \in \{16, 14, 12, 10, 8\}$ and $p \in \{true,false\}$. $\alpha$ stays at the default value of 4. 
 
\subsection{Contractable BOSS (cBOSS)~\cite{middlehurst2019scalable}}
\label{sec:cboss}

    Due to its grid-search and method of retaining ensemble members BOSS is unpredictable in its time and memory resource usage, and is impractical for larger problems. cBOSS significantly speeds up BOSS while retaining accuracy by improving how the transform parameter space is evaluated and the ensemble is formed. The main change from BOSS to cBOSS is that it utilises a filtered random selection of parameters to find its ensemble members. cBOSS allows the user to control the build through a time contract, defined as the maximum amount of time spent constructing the classification model.
        Algorithm~\ref{alg:cBOSS} describes the decision procedure for search and maintaining individual BOSS classifiers for cBOSS. 

    A new parameter $k$ (default 250) for the number of parameter combinations samples is introduced (line 7), of which the top $s$ with the highest accuracy are kept for the final ensemble (lines 13-19). The $k$ parameter is replaceable with a time limit $t$ through contracting. Each ensemble member is built on a subsample of the train data, (line 10) using random sampling without replacement of 70\% of the whole training data. An exponential weighting scheme for the predictions of the base classifiers is introduced, to produce a tilted distribution (line 18). 
    
    cBOSS was shown to be an order of magnitude faster than BOSS on both small and large datasets from the UCR archive while showing no significant difference in accuracy~\cite{middlehurst2019scalable}.

        \begin{algorithm}[ht]
        	\caption{cBOSS(A list of $n$ cases length $m$, ${\bf T}=({\bf X,y})$)}
        	\label{alg:cBOSS}
        	\begin{algorithmic}[1]
        \REQUIRE the number of parameter samples $k$, the max ensemble size $s$
\STATE Let $w$ be window length, $l$ be word length, $p$ be normalise/not normalise and $\alpha$ be  alphabet size.
\STATE Let ${\bf C}$ be a list of $s$ BOSS classifiers $({\bf c}_1,\ldots,{\bf c}_s)$
        		\STATE Let ${\bf E}$ be a list of $s$ classifier weights $({\bf e}_1,\ldots,{\bf e}_s)$
        		\STATE Let ${\bf R}$ be a set of possible BOSS parameter combinations
        		\STATE $i \leftarrow 0$
        		\STATE $lowest\_acc \leftarrow \infty, lowest\_acc\_idx \leftarrow \infty$
        		\WHILE {$i < k$ AND $|{\bf R}| > 0$}
        		    \STATE $[l,a,w,p] \leftarrow random\_sample({\bf R}) $
        		    \STATE $ {\bf R} = {\bf R} \setminus\{[l,a,w,p]\} $
        		
        		    \STATE ${\bf T'} \leftarrow$ subsample\_data(${\bf T}$)
        		    \STATE $cls \leftarrow$ baseBOSS(${\bf T'},l,a,w,p$)
        		    \STATE $acc \leftarrow$ LOOCV($cls$) \COMMENT{ {\em train data accuracy}}
        		    \IF{$i < s$}
        		        \IF{$acc < lowest\_acc$}
        		            \STATE $lowest\_acc \leftarrow acc$, $lowest\_acc\_idx \leftarrow i$
        		        \ENDIF
        		        \STATE $c_i \leftarrow cls$, $e_i \leftarrow acc^4$
        		    \ELSIF{$acc > lowest\_acc$}
        		        \STATE $c_{lowest\_acc\_idx} \leftarrow cls$, $e_{lowest\_acc\_idx} \leftarrow acc^4$
        		        \STATE $[lowest\_acc,lowest\_acc\_idx] \leftarrow$ find\_new\_lowest\_acc(${\bf C}$)
        		    \ENDIF
        		    \STATE $i \leftarrow i+1$
        		\ENDWHILE
        	\end{algorithmic}
        \end{algorithm}

\subsection{BOSS with Spatial Pyramids (S-BOSS)~\cite{large2019time}}
\label{sec:sboss}

BOSS intentionally ignores the locations of words in series, classifying based on the frequency of patterns rather than their location. For some datasets we know that the locations of certain discriminatory subsequences are important, however. Some patterns may gain importance only when in a particular location, or a mutually occurring word may be indicative of different classes depending on when it occurs. Spatial pyramids~\cite{lazebnik2006beyond} bring some temporal information back into the bag-of-words paradigm. 

\begin{figure}
	\centering
    \includegraphics[width=.9\linewidth,trim={7cm 2cm 4cm 2cm},clip]{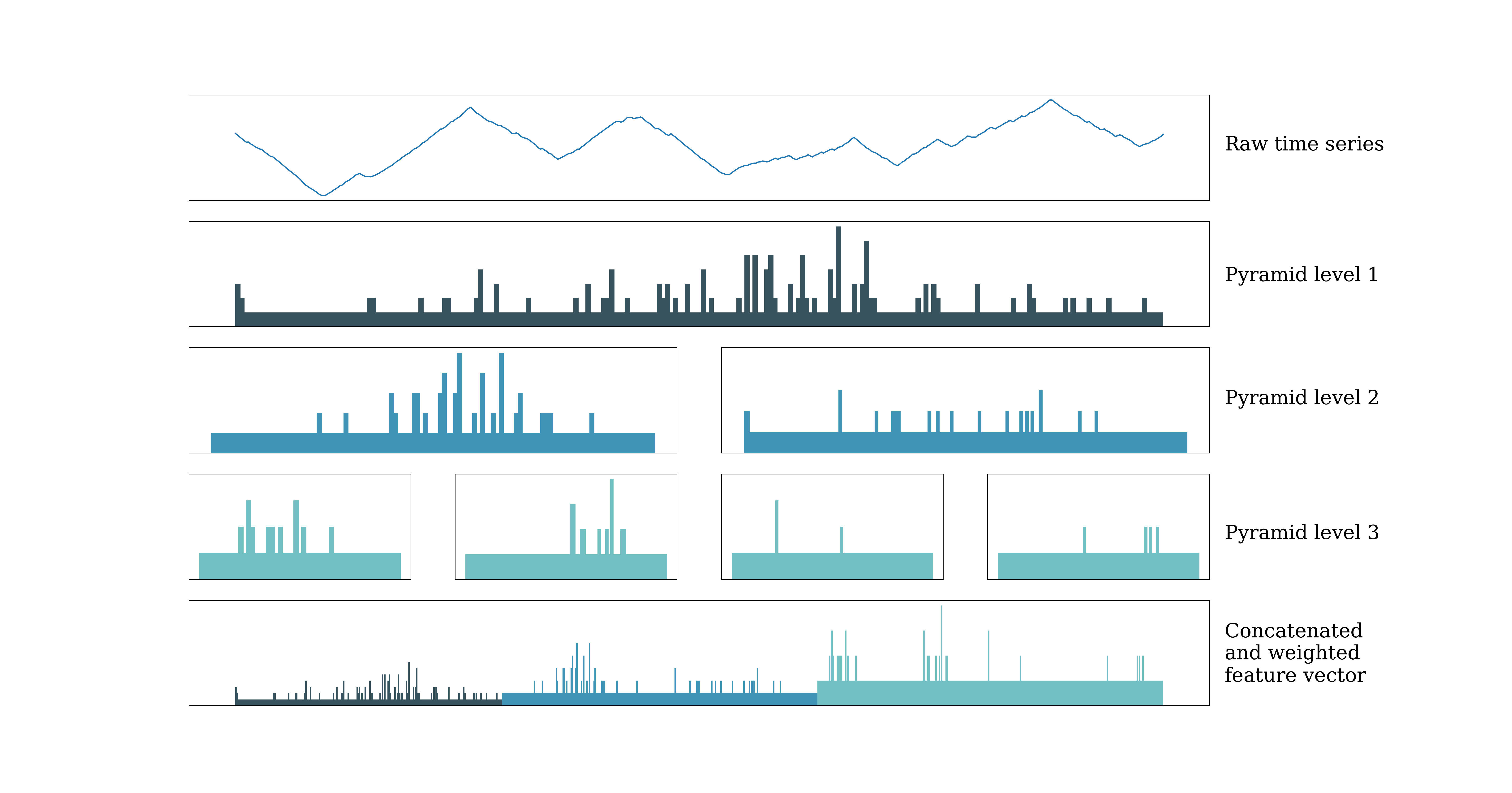}
    \caption{An example transformation of an OSULeaf instance to demonstrate the additional steps to form S-BOSS from BOSS. Note that each histogram is represented in a sparse manner; the set of words along the x-axis of each histogram at higher pyramid levels may not be equal. }
    \label{fig:s-boss}
\end{figure}

S-BOSS, described in Algorithm~\ref{alg:S-BOSS} and illustrated in figure~\ref{fig:s-boss}, incorporates the spatial pyramids technique into the BOSS algorithm. S-BOSS creates a standard BOSS transform at the global level (line 6), which constitutes the first level of the pyramid. An additional degree of optimisation is then performed to find the best pyramid height $h \in \{1,2,3\}$ (lines 11-16). Height defines the importance of localisation for this transform. Creating the next pyramid level involves creating additional histograms each sub-region of the series at the next scale. Histograms are weighted to give more importance to similarities in the same locations than global similarity, and are concatenated to form an elongated feature vector per instance. The histogram intersection distance measure, more commonly used for approaches using histograms, replaces the BOSS distance for the nearest neighbour classifiers. S-BOSS retains the BOSS ensemble strategy (line 17), such that each S-BOSS ensemble member is a BOSS transform with its own spatial pyramid optimisation plus nearest neighbour classifier. 

\begin{algorithm}[ht]
	\caption{S-BOSS(A list of $n$ cases length $m$, ${\bf T}=({\bf X,y})$)}
	\label{alg:S-BOSS}
	\begin{algorithmic}[1]
        \REQUIRE the set of possible $[a,w,p]$ parameter combinations ${\bf R}$, the set of possible $[l]$ parameter values ${\bf L}$, the maximum pyramid height $H$
        
        \STATE Let ${\bf C}$ be a list of $s$ BOSS classifiers $({\bf c}_1,\ldots,{\bf c}_s)$
		\FOR {$i \leftarrow  1$ to $|{\bf L}|$}
		    \STATE $bestAcc \leftarrow 0, bestCls \leftarrow \varnothing$ 
    		\FOR {$j \leftarrow  1$ to $|{\bf R}|$}
    		    \STATE $[a,w,p] \leftarrow {\bf R_j} $
    		    \STATE $cls \leftarrow$ baseBOSS(${\bf T},L_i,a,w,p$)
    		    \STATE $acc \leftarrow$ LOOCV($cls$) \COMMENT{{\em train data accuracy}}
    		    \IF{$acc > bestAcc$}
		            \STATE $bestAcc \leftarrow acc, bestCls \leftarrow cls$ 
		        \ENDIF
    		\ENDFOR
    		\STATE $cls \leftarrow bestCls$ 
    		\FOR {$h \leftarrow  1$ to $H$}
    		    \STATE $cls \leftarrow divideAndConcatenateBags(cls)$ 
    		    \STATE $acc \leftarrow$ LOOCV($cls$) \COMMENT{{\em train data accuracy}}
    		    \IF{$acc > bestAcc$}
		            \STATE $bestAcc \leftarrow acc, bestCls \leftarrow cls$ 
		        \ENDIF
    		\ENDFOR
    		
    		\STATE ${\bf C}_i \leftarrow bestCls$ 
		\ENDFOR
		\STATE $keepWithinBest({\bf C}, 0.92)$ \COMMENT{{\em keep those cls with train accuracy within 0.92 of the best}}
   \end{algorithmic}
\end{algorithm}

\subsection{Word Extraction for Time Series Classification (WEASEL)~\cite{schafer2017fast}}
\label{sec:weasel}
        \begin{algorithm}[ht]
    	\caption{WEASEL(A list of $n$ cases of length $m$, ${\bf T}=({\bf X,y})$)}
    	\label{alg:weasel}
    	\begin{algorithmic}[1]
    	\REQUIRE the word length $l$, the alphabet size $\alpha$, the maximal window length $w_{max}$, mean normalisation parameter $p$
    		\STATE Let ${\bf H}$ be the histogram ${\bf h}$
    		\STATE Let ${\bf B}$ be a matrix of $l$ by $\alpha$ breakpoints found by MCB using information gain binning
    		\FOR {$i \leftarrow  1$ to $n$}
    			\FOR {$w \leftarrow  2$ to $w_{max}$}
    				\FOR {$j \leftarrow 1$ to $m-w+1$}
    					\STATE ${\bf o}\leftarrow x_{i,j} \ldots x_{i,j+w-1}$
    					\STATE ${\bf q} \leftarrow$ DFT($o, w, p$) \COMMENT{ {\em {\bf q} is a vector of 	the complex DFT coefficients}}
    					\STATE ${\bf q'} \leftarrow$ ANOVA-F($q, l, y$) \COMMENT{ {\em use only the {\bf l} most discriminative ones}}					
    					\STATE ${\bf r} \leftarrow$ SFAlookup(${\bf q', B}$)
    					\STATE $pos \leftarrow $index(${\bf w, r}$)
    					\STATE ${h}_{i,pos} \leftarrow {h}_{i,pos} + 1$
    				\ENDFOR					
    			\ENDFOR
    		\ENDFOR		
    		\STATE $h \leftarrow \chi^2(h, y)$ \COMMENT{ {\em feature selection using the chi-squared test} }
    		\STATE fitLogistic($h, y$)
    	\end{algorithmic}
        \end{algorithm}

    Like BOSS, WEASEL performs a Fourier transform on each window, creates words by discretisation, and forms histograms of words counts. It also does this for a range of window sizes and word lengths. However, there are important differences. WEASEL is not an ensemble nearest neighbour classifiers. Instead, WEASEL constructs a single feature space from concatenated histograms for different parameter values, then uses logistic regression and feature selection. Histograms of individual words and bigrams of the previous non-overlapping window for each word are used. Fourier terms are selected for retention by the application of an F-test. The retained values are then discretised into words using information gain binning (IGB), similar to the MCB step in BOSS. The number of features is further reduced using a chi-squared test after the histograms for each instance are created, removing any words which score below a threshold. It performs a parameter search for $p$ (whether to normalise or not) and over a reduced range of $l$, using a 10-fold cross-validation to determine the performance of each set. The alphabet size  $\alpha$ is fixed to 4 and the $chi$ parameter is fixed to 2. Algorithm~\ref{alg:weasel} gives an overview of WEASEL, although the formation and addition of bigrams is omitted for clarity.

\section{Temporal Dictionary Ensemble (TDE)}
\label{sec:tde}
    The easiest algorithms to combine are cBOSS and S-BOSS. cBOSS speeds up BOSS through subsampling training cases and random parameter selection.
    The number of levels parameter introduced by S-BOSS can be included in the random parameter selection used by cBOSS. For comparisons we call this naive hybrid of algorithms cS-BOSS. We use this as a baseline to justify the point of the extra complexity we introduce in TDE.  
    Algorithm~\ref{alg:TDE} provides an overview of the ensemble build process for TDE, which follows the general structure and weighting scheme of cBOSS. The classifier returned by improvedBaseBOSS includes spatial pyramids and also includes the following enhancements taken from both S-BOSS and WEASEL. Like S-BOSS, it uses the histogram intersection distance measure which has been shown to be more accurate than BOSS distance~\cite{large2019time}. It uses bigram frequencies in the same way as WEASEL. Base classifiers can use either IGB from WEASEL or MCB from BOSS in the discretisation. TDE samples parameters from the range given in Table~\ref{tab:parameters} using a method sampleParameters.

\begin{table}[ht]
    \centering
    \caption{Parameter ranges for TDE base classifier selection.}
    \begin{tabular}{l|l}
    
    \hline
    Parameter       & Range \\ 
    \hline
    Word lengths    & $l=\{16,14,12,10,8\}$\\
    Window lengths  & $w=\{10...m\}$  \\
    Normalise       & $p =\{true,false\}$ \\
    No.pyramid levels  & $h =\{1,2,3\}$ \\
    Discretisation        &    $b =\{MCB,IGB\}$ \\
    \hline
    
    \end{tabular}
    \label{tab:parameters}
\end{table}

        \begin{algorithm}[ht]
        	\caption{TDE(A list of $n$ cases length $m$, ${\bf T}=({\bf X,y})$)}
        	\label{alg:TDE}
        	\begin{algorithmic}[1]
                \REQUIRE the number of parameter samples $k$, the max ensemble size $s$
                \STATE Let $w$ be window length, $l$ be word length, $p$ be normalise/not normalise, $\alpha$ be alphabet size, $h$ be number of pyramid levels and $b$ be MCB or IGB discretisation.
                \STATE Let ${\bf C}$ be a list of $s$ BOSS classifiers $({\bf c}_1,\ldots,{\bf c}_s)$
        		\STATE Let ${\bf E}$ be a list of $s$ classifier weights $({\bf e}_1,\ldots,{\bf e}_s)$
        		\STATE Let ${\bf G}$ be a list of $k$ BOSS parameter and accuracy pairs $({\bf g}_1,\ldots,{\bf g}_k)$
        		\STATE Let ${\bf R}$ be a set of possible BOSS parameter combinations
        		\STATE $i \leftarrow 0$
        		\STATE $lowest\_acc \leftarrow \infty, lowest\_acc\_idx \leftarrow \infty$
        		\WHILE {$i < k$ AND $|{\bf R}| > 0$}
            		\STATE $[l,a,w,p,h,b] \leftarrow$ chooseParameters(${\bf R}$,${\bf G}$,$i$)
        		    \STATE $ {\bf R} = {\bf R} \setminus\{[l,a,w,p,h,b]\} $
        		    \STATE ${\bf T'} \leftarrow$ subsampleData(${\bf T}$)
        		    \STATE $cls \leftarrow$ improvedBaseBOSS(${\bf T'},l,a,w,p,h,b$)
        		    \STATE $acc \leftarrow$ LOOCV($cls$) \COMMENT{ {\em train data accuracy}}
        		    \IF{$i < s$}
        		        \IF{$acc < lowest\_acc$}
        		            \STATE $lowest\_acc \leftarrow acc$, $lowest\_acc\_idx \leftarrow i$
        		        \ENDIF
        		        \STATE $c_i \leftarrow cls$, $e_i \leftarrow acc^4$
        		    \ELSIF{$acc > lowest\_acc$}
        		        \STATE $c_{lowest\_acc\_idx} \leftarrow cls$, $e_{lowest\_acc\_idx} \leftarrow acc^4$
        		        \STATE $[lowest\_acc,lowest\_acc\_idx] \leftarrow$ findNewLowestAcc(${\bf C}$)
        		    \ENDIF
        		    \STATE $g_i \leftarrow \{[l,a,w,p,h,b],acc\}$
        		    \STATE $i \leftarrow i+1$
        		\ENDWHILE
        	\end{algorithmic}
        \end{algorithm}
    
\subsection{Gaussian process parameter selection}

    The increase in the parameter search space caused by the inclusion of pyramids and IGB parameters makes the random  parameter selection used by cBOSS less effective. Instead, TDE uses a guided parameter selection for ensemble members  inspired by Bayesian optimisation~\cite{snoek2012practical}. A Gaussian process model is built over the regressor parameter space ${\bf R}$ for parameters $[l,a,w,p,h,b]$ to predict the accuracy, using the previously observed (parameter, accuracy) pairs ${\bf G}$.   
    
    A Gaussian Process~\cite{williams2006gaussian} describes a distribution over functions, $f(\vec{x}) \sim \mathcal{GP}(m(\vec{x}, k(\vec{x},\vec{x}')))$, characterised by a mean function, $m(\vec{x})$, and a covariance function, $k(\vec{x},\vec{x}')$, such that
    \begin{eqnarray*}
       m(\vec{x})          & = & \mathbb{E}\left[f(\vec{x})\right], \\
       k(\vec{x},\vec{x}') & = & \mathbb{E}\left[(f(\vec{x}) - m(\vec{x}))(f(\vec{x}') - m(\vec{x}'))\right],
    \end{eqnarray*}
    where any finite collection of values has a joint Gaussian distribution.  Commonly the mean function is
    constant, $m(\vec{x}) = \gamma$, or even zero, $m(\vec{x}) = 0$.  The covariance function $k(\vec{x},\vec{x}')$ encodes the expected similarity of the function evaluated at pairs of input-space
    vectors, $\vec{x}$ and $\vec{x}'$.  For example, the squared exponential covariance function,
    \begin{displaymath}
        k(\vec{x}, \vec{x}') = \sigma_f^2\exp\left\{-\frac{\left(\vec{x} - \vec{x}'\right)^2}{2\ell^2}\right\},
    \end{displaymath}
    encodes a preference for smooth functions, where $\ell$ is a hyper-parameter that specifies the characteristic length-scale of the covariance functions (large values yield smoother functions) and
    $\sigma_f$ governs the magnitude of the variance.  
    
    Typically in a regression setting the response variable of the training samples, $\mathcal{D} = \left\{(\vec{x}_i, y_i)~|~i = 1, 2, \ldots, n\right\}$, are assumed to be realisations of a deterministic function that have been corrupted by additive Gaussian noise, i.e.
    \begin{displaymath}
       y_i = f(\vec{x}_i) + \varepsilon_i,
       \qquad
       \mathrm{where}
       \qquad
       \varepsilon_i \sim \mathcal{N}\left(0, \sigma_n^2\right).
    \end{displaymath}
    In that case, the joint distribution of the training sample, and a single test point, $\vec{x}_\ast$, is given by,
    \begin{displaymath}
       \left[\begin{array}{c}
                \vec{y} \\
                f_\ast  
             \end{array}\right] \sim \mathcal{N}\left(\vec{0}, \left[
                \begin{array}{cc}
                \Matrix{K} + \sigma_n^2\Matrix{I} & \vec{k}_\ast \\
                \vec{k}_\ast^T & k(\vec{x}_\ast, \vec{x}_\ast) 
                \end{array}
                \right]\right),
    \end{displaymath}
    where $\Matrix{K}$ is the matrix of pairwise evaluation of the covariance function for all points belonging to the training sample and $\vec{k}_\ast$ is a column vector of the evaluation of the covariance function for the test point and each of the training points.  The Gaussian predictive distribution for the test point is then specified by
    \begin{eqnarray*}
       \bar{f}_\ast & = & \vec{k}_\ast^T\left(\Matrix{K} + \sigma_n^2\Matrix{I}\right)^{-1}\vec{y},\\
       \mathbb{V}\left[ f_\ast \right] & = & k(\vec{x}_\ast, \vec{x}_\ast) - \vec{k}_\ast^T\left(\Matrix{K} + \sigma_n^2\Matrix{I}\right)^{-1}\vec{k}_\ast.
    \end{eqnarray*}
    
    The hyper-parameters of the Gaussian process can be handled by tuning them, often via maximisation of the marginal likelihood, or by full Bayesian marginalisation, using an appropriate hyper-prior distribution. For further details, see Williams and Rasmussen~\cite{williams2006gaussian}. We use a basic form of GP and treat all the regressors (TDE parameters) as continuous. The bestPredictedParameters operation in line 5 of Algorithm~\ref{alg:gp} is limited to the same parameter ranges used for random search given in Table~\ref{tab:parameters}. 
 
        \begin{algorithm}
        	\caption{chooseParameters(${\bf R}$,${\bf G}$,$i$)}
        	\label{alg:gp}
        	\begin{algorithmic}[1]
                    \IF{$i < 50$}
        		        \STATE $[l,a,w,p,h,b] \leftarrow$ randomSample(${\bf R}$)
        		    \ELSE
        		        \STATE $gp \leftarrow$ buildGaussianProcesses(${\bf G}$)
        		        \STATE $[l,a,w,p,h,b] \leftarrow$ bestPredictedParameters(${\bf R}, gp$)
        		        \ENDIF 
                    \RETURN $[l,a,w,p,h,b]$ 
        	\end{algorithmic}
        \end{algorithm}

\section{Results} 
\label{sec:results}

    Our experiments are run on 112 datasets from the recently expanded UCR/UEA archive~\cite{dau2019ucr}, removing any datasets that are unequal length or contain missing values. We also remove the dataset Fungi as it only provides a single train case for each class. For each classifier dataset combination we run 30 stratified resamples, with the first sample being the original train test split.
    For reproducability each dataset resample and classifier is seeded to its resample number. 
    All experiments were run single threaded on a high performance computing cluster with a run time limit of 7 days. We used an open source Weka compatible code base called tsml for experimentation\footnote{\url{https://github.com/uea-machine-learning/tsml}}. Implementations of BOSS, cBOSS, S-BOSS, WEASEL, HIVE-COTE and TS-CHIEF provided by the algorithm inventors are all available in tsml. InceptionTime and ROCKET experiments were run using the Python based package sktime and a deep learning extension thereof\footnote{\url{https://github.com/sktime}}.

Guidance on how to recreate the resamples and code to reproduce the results is available on the accompanying website\footnote{\url{https://sites.google.com/view/ecmlpkdd-tde/home}}. We also provide results for all classifiers used in experimentation.  
        
Our experiments test are designed to test whether TDE is better in terms of predictive performance and run time than other dictionary based classifiers, and whether it improves HIVE-COTE when it replaces BOSS in the meta ensemble HIVE-COTE. 

\subsection{TDE vs other dictionary classifiers}

For the dictionary classifiers, we were only able to obtain complete results for 107 of the 112 datasets. This was due to the long run time of S-BOSS and WEASEL. For consistency, in this Section we only present results with these datasets. The missing problems are: ElectricDevices; FordA; FordB; HandOutlines; and NonInvasiveFetalECGThorax2.

\begin{figure}[t]
	\centering
    \includegraphics[width=\linewidth,trim={0cm 8cm 0cm 5cm},clip]{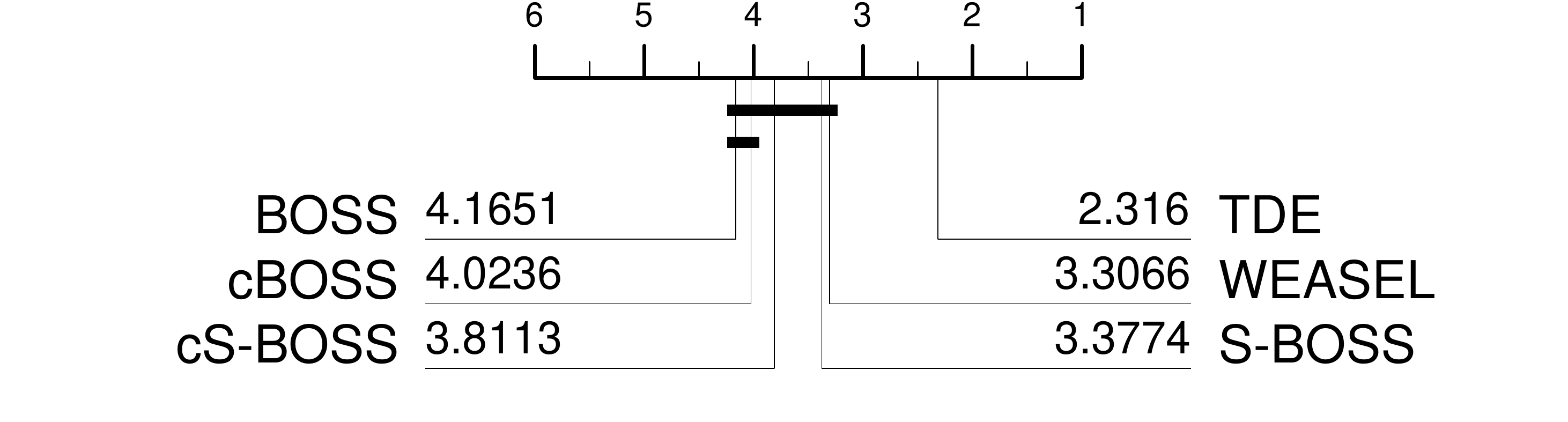}
    \caption{Critical difference diagram for six dictionary based classifiers on 107 UCR time series classification problems. Full results are available on the accompanying website.}
    \label{fig:dictionary}
\end{figure}

Figure~\ref{fig:dictionary} shows a a critical difference diagram~\cite{demvsar2006statistical} for the six dictionary based classifiers considered. 
The number on each line is the average rank of an algorithm over 107 UCR datasets (lower is better).  The solid bars are cliques. There is no detectable significant difference between classifiers in the same clique. 
Comparison of the performance of classifiers is done using pairwise Wilcoxon signed rank tests and cliques are formed using the Holm correction, following recommendations from \cite{benavoli2016should} and \cite{garcia2008extension}. 

TDE is significantly more accurate than all other classifiers. There are then two cliques: BOSS and cBOSS are significantly worse than S-BOSS, cS-BOSS and WEASEL. This also confirms that the simple hybrid cS-BOSS is no better than S-BOSS in terms of accuracy.

Table~\ref{tab:runtime} summarises the run time of these classifiers. All the classifiers can complete on most of the UCR problems in minutes. S-BOSS is the slowest algorithm. 
Four problems are not included in this study because either S-BOSS or WEASEL did not to finish within seven days. For WEASEL this was caused by the requirement to cross validate to estimate the accuracy for HIVE-COTE. S-BOSS is just relatively slow. We omit these four problems from all analysis for completeness. All experiments are conducted sequentially on a single processor. TDE completes all the problems in under a day. It is faster than WEASEL and considerably faster than S-BOSS. 

\begin{table}[b]
    \centering
    \caption{Summary of run time for six dictionary based classifiers over 107 UCR problems. The median time over 30 resamples is used for each dataset.}
    \label{tab:runtime}
    \begin{tabular}{l|c|c}
    
    \hline
    Classifier & Max run time (hrs) & Total run time (hrs)  \\ 
    \hline
    BOSS       &  11.33   & 52.10           \\
    cBOSS      &  0.63  &  3.74                \\
    S-BOSS     &  34.11   & 148.82                    \\
    cS-BOSS    &  2.91   & 12.62                   \\
    WEASEL     &  4.86   & 28.50                  \\
    TDE        &  5.67   & 21.73                     \\ 
    \hline
    
    \end{tabular}
\end{table}

\begin{figure}
	\centering
    \includegraphics[width=0.9\linewidth,trim={0cm 0cm 0cm 0cm},clip]{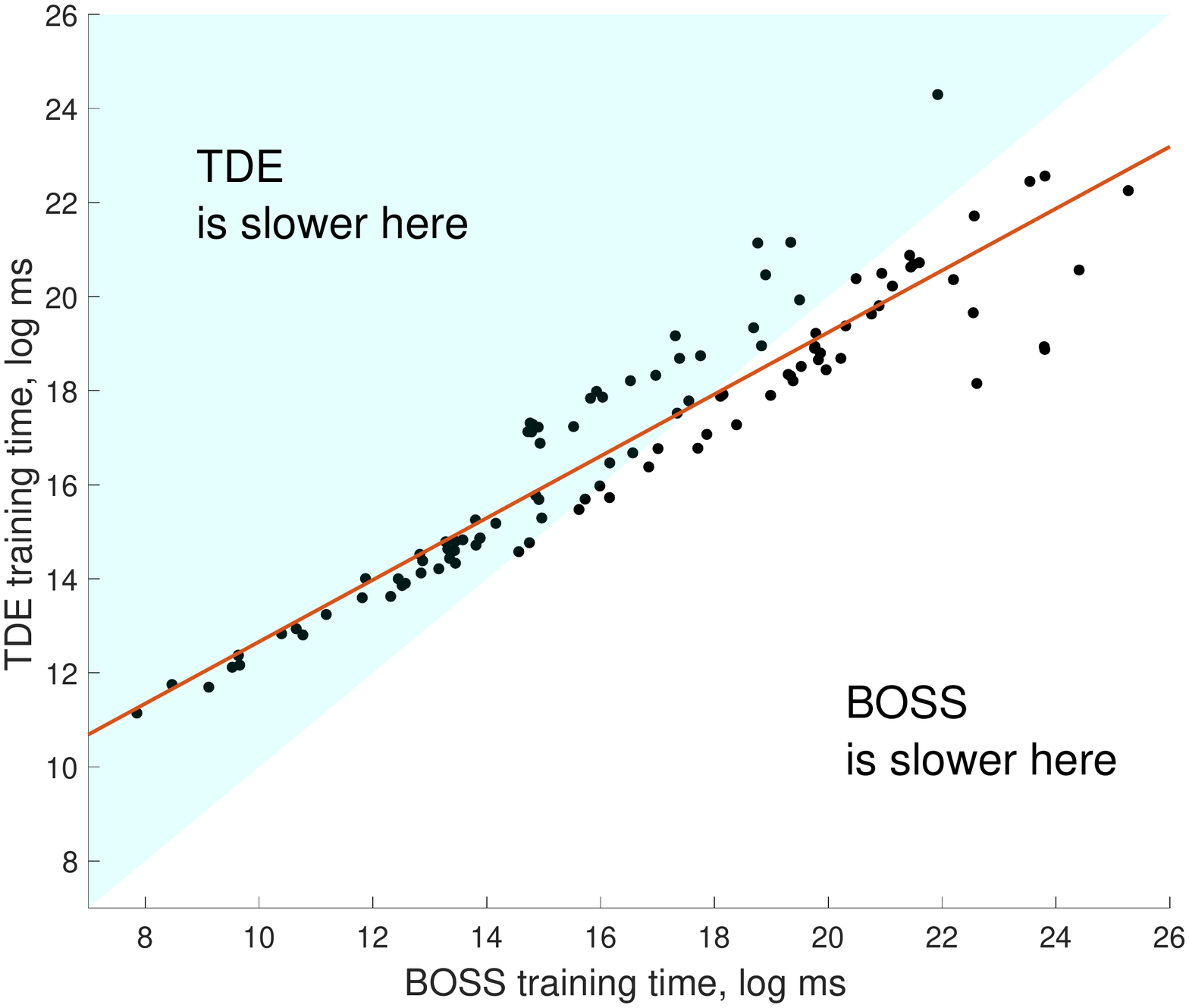}
    \caption{Pairwise scatter diagram, on log scale, of TDE and BOSS training times. TDE has larger overheads which make it slower on smaller problems, but it scales much better towards larger problems. } 
    \label{fig:timing}
\end{figure}

The max run timing for BOSS and S-BOSS demonstrate the problem with the traditional BOSS algorithm addressed by cBOSS and cS-BOSS: the ensemble design means they can have a long runtime and it is not very predictable when this will happen. Figure~\ref{fig:timing} shows the scatter plot of runtime for BOSS vs TDE and demonstrates that TDE scales much better than BOSS.

\subsection{TDE with HIVE-COTE}

TDE is significantly more accurate than all other dictionary based time series classification algorithms, and faster than the best of the rest, S-BOSS and WEASEL. We believe there is merit in finding the best single representation classifier because there will be occasions when domain knowledge would recommend a single approach. However, with no domain knowledge, the state of the art in time series classification involves hybrids built on multiple representations, or deep learning to fit a bespoke representation. HIVE-COTE is a meta ensemble of classifiers built using different representations.

All HIVE-COTE variants used in our experiments are built with four components: the Random Interval Spectral Ensemble (RISE), Shapelet Transform Classifier (STC) and Time Series Forest (TSF) plus one other classifier (see~\cite{lines2018time} for details). We omit the Elastic Ensemble because it is infeasible to run it on a large number of problems. The other components are relatively fast and can complete a single resample of one problem in under a day for all problems. The latest version of STC on the open source repository tsml does not do a full enumeration of the shapelet space. Instead, it randomly samples shapelets for a fixed time. We set this to 4 hours for all experiments with STC. 

With these other settings fixed, we have reconstructed HIVE-COTE using TDE instead of BOSS. We call this HC-TDE for differentiation purposes.

\begin{figure}[b]
	\centering
    \includegraphics[width=\linewidth,trim={0cm 9cm 0cm 5cm},clip]{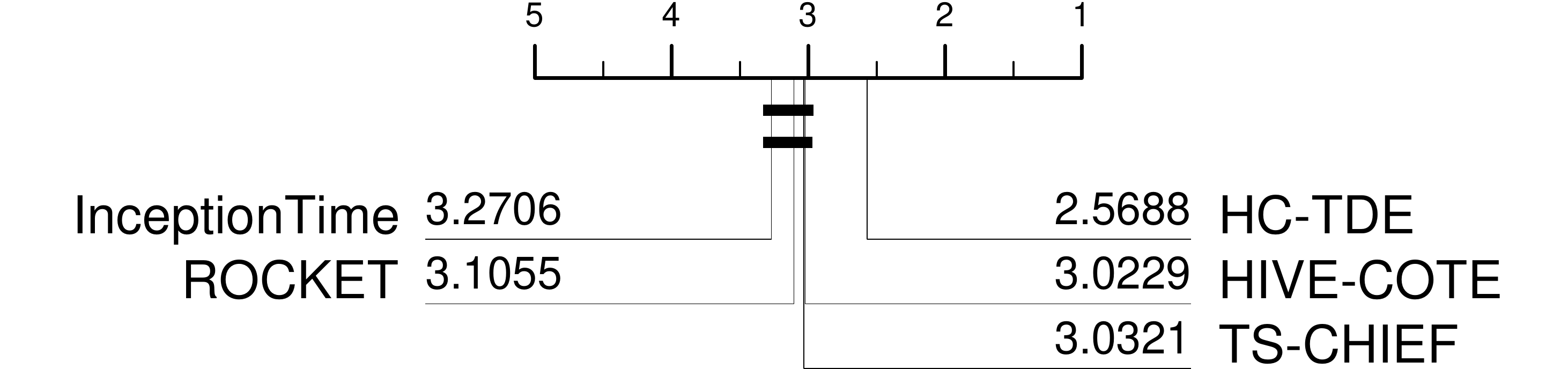}
    \caption{Critical difference diagram for six dictionary based classifiers on 109 UCR time series classification problems. Full results are available on the accompanying website.}
    \label{fig:sota}
\end{figure}

TS-CHIEF~\cite{shifaz2020ts} is a tree ensemble that embeds dictionary, spectral and distance based representations, and is set to build 500 trees. InceptionTime~\cite{fawaz2019inceptiontime} is a deep learning ensemble, combining 5 homogeneous networks each with random weight initialisations for stability. ROCKET~\cite{dempster2019rocket} uses a large number, 10,000, of randomly parameterised convolution kernels in conjunction with a linear ridge regression classifier. We use the configurations of each classifier described in their respective publications.

Figure~\ref{fig:sota} shows the ranked performance of HC-TDE against HIVE-COTE, TS-CHIEF, InceptionTime and ROCKET on 109 problems. We are missing three data, HandOutlines, NonInvasiveFetalECGThorax1 and NonInvasiveFetalECGThorax2 because TS-CHIEF could not complete them within the seven day limit. 

\begin{figure}[t]
	\centering
    \includegraphics[width=0.9\linewidth,trim={0cm 0cm 0cm 0cm},clip]{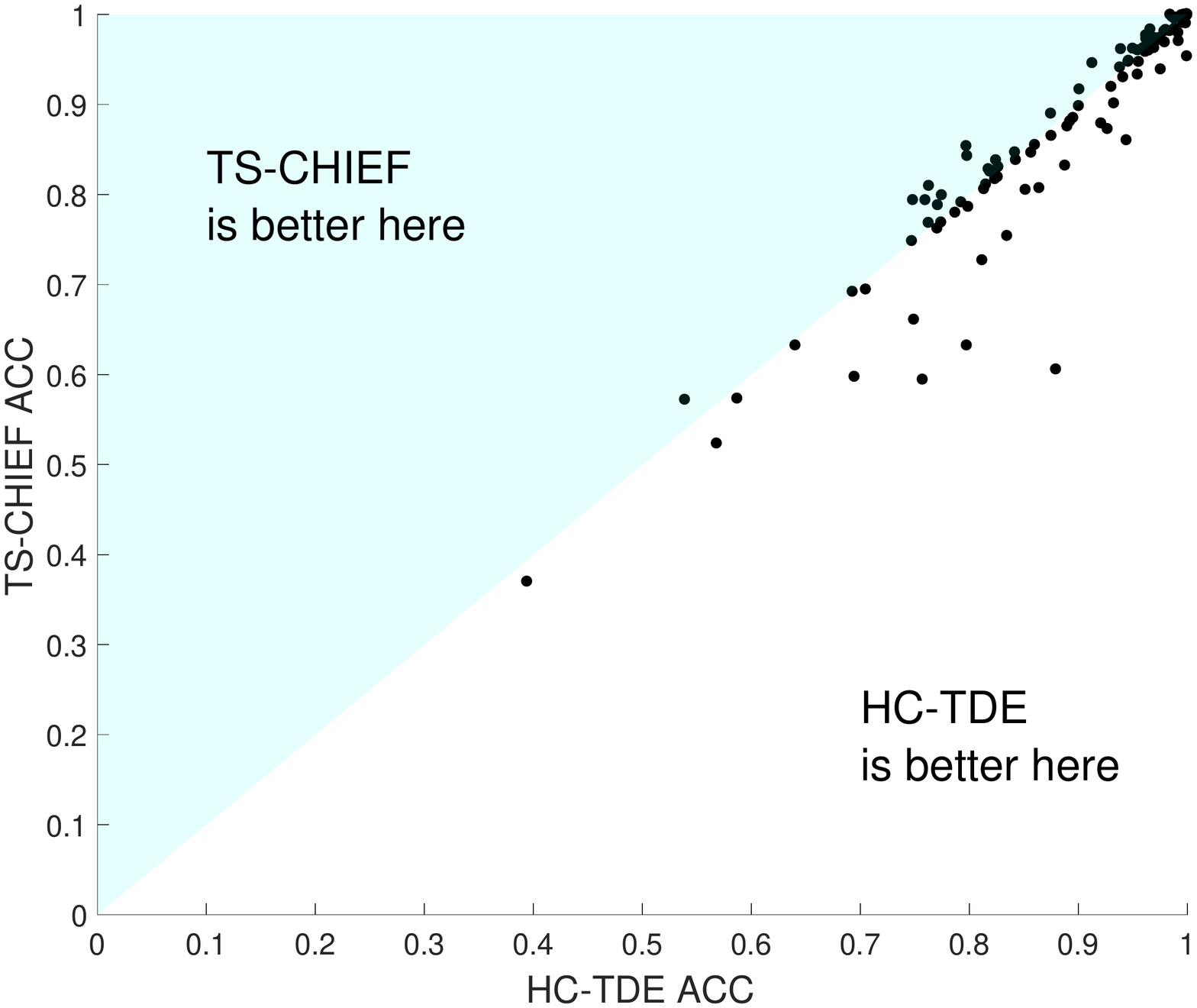}
    \caption{Scatter plot of TS-CHIEF against HC-TDE. HC-TDE wins on 62, draws on 6 and loses on 41 data sets.}
    \label{fig:scatter}
\end{figure}
HC-TDE is significantly better than all four algorithms currently considered state of the art. The actual differences between HIVE-COTE and HC-TDE are understandably small, given their similarities. However, they are consistent: replacing BOSS with TDE improves HIVE-COTE on 69 problems, and makes it worse on just 32 (with 8 ties). HC-TDE does show significant variation to TS-CHIEF (see Figure~\ref{fig:scatter}) and is on average over 1\% more accurate. HC-TDE is the top performing algorithm using a range of performance measures such as AUROC, F1 and balanced accuracy (see accompanying website). The improvement over InceptionTime is even greater: it is on average 1.5\% more accurate.

\begin{figure}[t]
	\centering
    \includegraphics[width=\linewidth,trim={0cm 6cm 0cm 5cm},clip]{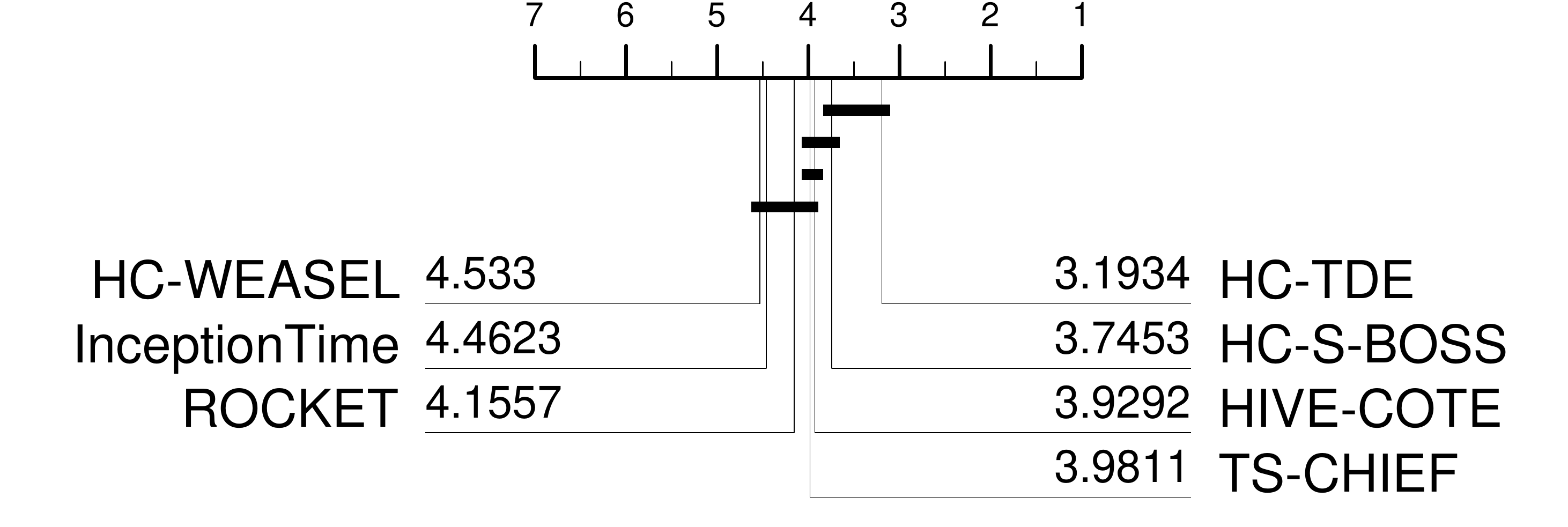}
    \caption{Critical difference diagram for seven classifiers on 106 UCR time series classification problems. Full results are available on the accompanying website.}
    \label{fig:hc}
\end{figure}

It is worth considering whether replacing BOSS with either S-BOSS or WEASEL would give as much improvement to HIVE-COTE as TDE does. We replaced BOSS with WEASEL (HC-WEASEL) and S-BOSS (HC-S-BOSS).
Figure~\ref{fig:hc} shows the performance of these relative to HC-TDE, InceptionTime and TS-CHIEF. Whilst it is true that HC-S-BOSS is not significantly worse than HC-TDE, it is also not significantly better than the current state of the art. HC-WEASEL does not perform well. We speculate that this is because the major differences in WEASEL mean that its improvement is at problems better suited to other representations, and this improvement comes at the cost of worse performance at problems suited to dictionary classifiers.  

\section{Conclusion}
\label{sec:conclusion}
TDE combines the best elements of existing dictionary based classifiers with a novel method of improving the ensemble through a Gaussian process model for parameter selection. TDE is more accurate and scalable than current top performing dictionary algorithms. When we replace BOSS with TDE in HIVE-COTE, the resulting classifier is significantly more accurate than the current state of the art. TDE has some drawbacks. It is memory intensive. It requires about three times more memory than BOSS, and the maximum memory required was 10 GB for ElectricDevices.  Like all nearest neighbour classifiers, TDE is relatively slow to classify new cases. If fast predictions are required, WEASEL may be preferable. Future work will focus on making TDE more scalable.

\section*{Acknowledgements}{
     This work is supported by the UK Engineering and Physical Sciences Research Council (EPSRC) iCASE award T206188 sponsored by British Telecom. The experiments were carried out on the High Performance Computing Cluster supported by the Research and Specialist Computing Support service at the University of East Anglia.
}

\bibliographystyle{splncs04}
\bibliography{paper}

\end{document}